\documentclass[11pt,a4paper]{article}

\usepackage{float}
\usepackage{url}
\usepackage{amsmath,amsfonts}
\usepackage{algorithmic}
\usepackage{array}
\usepackage{graphicx}
\usepackage[hidelinks]{hyperref}

\title{\bfseries
Constrained Hebbian Learning Supports Efficient Representational Allocation under Structural Constraints
}

\date{July 2026}

\author{
Patrick Inoue$^{*\dag}$ \and
Florian Röhrbein$^{\dag}$ \and
Andreas Knoblauch$^{*}$ \\[1ex]
{\small $^{*}$KEIM Institute, Albstadt-Sigmaringen University, Germany}\\
{\small $^{\dag}$Department of Computer Science, Chemnitz University of Technology, Germany}
}

\begin{document}

\maketitle

\begin{abstract}

\noindent \textbf{Introduction:} Biological systems are constrained by anatomy and metabolism, with synaptic maintenance consuming substantial energy and spatial limitations restricting connectivity. These constraints favor neural codes that compress behaviorally relevant information into low-redundancy patterns. Here, we investigate whether a strictly local, excitatory competitive Hebbian rule can implement synaptic resource allocation under anatomical and metabolic constraints, and whether the resulting representations occupy a more favorable cost-performance regime than reference learning rules.

\noindent \textbf{Methods:} Representational cost is operationalized using mutual-information-based quantities derived from the Variational Information Bottleneck objective. Experiments employ three audiovisual benchmarks (AVE, Kinetics-Sounds, VGGSound100) with fixed upstream audiovisual embeddings as a controlled higher-level sensory regime. We isolate downstream associative plasticity and compare Hebbian learning against biologically inspired local target-propagation (DDTP) and backpropagation (BP) under matched sparsity and architectural constraints.

\noindent \textbf{Results:} Hebbian learning yields lower task-information cost (CTI) than sparse BP- and DDTP-trained networks in the main compressed comparisons, while reaching a CTI range comparable to shallow nonnegativity-constrained BP. Rather than uniformly improving classification performance, Hebbian learning shifts the trade-off between task-relevant information and representational cost, yielding lower CTI at comparable functional performance in several settings.

\noindent \textbf{Discussion:} Overall, the results indicate an explicit cost-performance trade-off rather than uniform accuracy gains, while maintaining functional task performance under the evaluated constraints. For a given level of task-relevant information, less information about the input is retained; task performance remains functional, although it can be slightly reduced on some datasets. These results support a computational-neuroscientific interpretation of Hebbian learning as a local mechanism for synaptic resource allocation, rather than as a general-purpose strategy for maximizing audiovisual classification accuracy.

\end{abstract}

\section{Introduction}
\label{introduction}

\noindent Information is physical: each neural computation consumes metabolic energy \cite{sengupta}. A substantial fraction of the brain's energy consumption arises from maintaining functional synapses \cite{knoblauch, attwell2001energy, lennie2003}. In addition to synaptic and signaling costs, neurons incur baseline metabolic expenses to sustain ionic gradients. These energetic demands, together with spatial and anatomical constraints that prevent fully connected networks and limit both the total number of synapses and the wiring density within cortical circuits, impose strict restrictions on how information can be represented and transmitted. Additional limitations arise from the restricted availability of metabolic substrates, such as glucose and oxygen \cite{Buxton2010,Buxton2021}, further constraining neural coding strategies.

Therefore, neurons do not encode all sensory details. Instead, they prioritize the features most relevant to behavior and cognition. For example, neurons in the primary and secondary visual cortices respond selectively to edges, orientations, and other behaviorally significant visual features, capturing the structure of the scene without representing every pixel \cite{hubel1968receptive, geisler2001edge}. This exemplifies an efficient coding principle, whereby the same behaviorally relevant information can be captured with sparser, lower-dimensional activity patterns and, together with structural plasticity, can be represented by a reduced set of functional synapses \cite{barlow1961possible, olshausen1996emergence}. 

Each activity pattern, defined as a specific combination of co-activated neurons that represent a stimulus, memory, or association, is instantiated through a particular configuration of functional synapses \cite{knoblauch2010memory}. Networks encoding a larger number of distinct patterns require more functional synapses, thereby increasing baseline metabolic expenditure \cite{sengupta, attwell2001energy,lennie2003}. By confining activity to lower-dimensional subspaces or exploiting redundancy, neural codes achieve the same informational content with fewer synaptic states, thus reducing energy expenditure. Hence, the dimensionality, pattern diversity, and redundancy of neural representations determine the metabolic energy required to sustain functional network states.

These considerations are not purely theoretical, as the energetic costs of maintaining neuronal activity patterns are grounded in the physical instantiation of neural information, with each distinct activity pattern corresponding to a configuration of ionic fluxes, membrane potentials, and synaptic states.  \cite{sengupta} provide a formal link between physical work and the information encoded in these hidden states, showing that reducing uncertainty about microscopic states requires energy expenditure and that sustaining ordered low-entropy configurations increases metabolic cost \cite{helmholtz1882}. Conceptually, the diversity of activity patterns can be quantified using representational entropy \cite{li2023energy}. Higher entropy reflects a larger number of distinct hidden synaptic states, which, according to \cite{sengupta}, increases baseline metabolic expenditure. Compressing information into structured, low-entropy codes is therefore expected to lower energy requirements by stabilizing low-energy firing patterns and limiting unnecessary spikes.

Directly estimating the entropy for natural sensory input is intractable. The Free-Energy Principle (FEP) addresses this problem by introducing variational free energy as a tractable upper bound on sensory surprise \cite{friston2010free, fristonfree}. Following \cite{fristonfree}, who connects variational free energy minimization to the Information Bottleneck (IB), we adopt these principles as a normative reference to evaluate representational efficiency. Computationally, we implement this perspective with the Variational Information Bottleneck (VIB) \cite{alemi}, which provides a tractable method to quantify task-relevant latent codes while trading off predictive accuracy and compressibility. 

By formalizing representational efficiency in these terms, the framework produces biologically grounded metrics to compare ANNs with respect to representational cost under a biologically motivated proxy and the compactness of task-relevant representations. These metrics reflect constraints that govern cortical networks, which operate under strict metabolic and anatomical limitations \cite{Buxton2021, knoblauch2017, fristonfree}. In contrast, ANNs are typically optimized for hardware-oriented objectives such as throughput, numerical precision and memory access. On conventional computing platforms, the theoretical metabolic cost of maintaining synapses or activity patterns is irrelevant, because hardware such as CPUs and GPUs does not incur energy expenditures analogous to neural tissue. Redundant or inactive weights do not reduce computational cost in practice, since conventional hardware executes all operations regardless of synaptic strength. Consequently, networks can be scaled to very large widths and depths without adhering to the metabolic or anatomical constraints observed in the brain, frequently resulting in overparameterization and connectivity that would be metabolically inefficient \cite{frankle2019lottery, chang2020provable}.

To mitigate the symptomatic consequences of overparameterization, training heuristics such as dropout, skip connections, and batch normalization are commonly employed. However, these techniques primarily address symptomatic issues, such as unstable gradients and overfitting, and do not remove the underlying redundancy. Explicit interventions such as pruning or sparsity-inducing penalties can remove redundant parameters. Empirical and theoretical work has already demonstrated this, exemplified by canonical architectures including AlexNet and VGG--16, showing that large fractions of parameters can be removed with negligible loss in accuracy \cite{frankle2019lottery, chang2020provable,han2015learning, li2016pruning}. Yet these pruning methods typically rely on global network statistics and generally ignore biological constraints, such as the locality of plasticity. Consequently, most learning algorithms, including widely used conventional architectures and many implementations of biologically inspired rules, are primarily optimized for execution on standard computing hardware, which limits their ability to provide mechanistic insight into synaptic resource allocation under anatomically and metabolically relevant constraints.

This divergence exposes a critical research gap, as conventional ANNs are not routinely evaluated or optimized with mechanisms that respect the local, metabolically constrained principles of neuronal resource assignment. To probe this gap, we employ biologically motivated latent-code metrics as a methodological tool to quantify synaptic resource utilization. In this context, we also apply an excitatory Hebbian learning rule \cite{neucomp, bwhpc, ijcnn}, which generates sparse, nonnegative synaptic weights consistent with Dale's law \cite{dale} and implements implicit neural principal component analysis (PCA) through activity-dependent competition, thereby approximating low-dimensional, decorrelated task-relevant representations. This framework enables us to move beyond descriptive evaluation toward mechanistic hypotheses for how neurons might allocate synaptic resources efficiently in a biologically plausible manner, grounded in known correlation-based plasticity mechanisms. Importantly, these datasets are used here not as end-to-end performance benchmarks, but as controlled assays for testing how different learning rules allocate representational resources across unimodal and bimodal inputs under matched architectural settings.

We hypothesize that these properties enable the Hebbian rule to learn compact, task-relevant representations under constraints inspired by anatomical structure and metabolic considerations. To test this hypothesis, we compare its performance against biologically inspired mechanisms, such as Dense Difference Target Propagation (DDTP) \cite{meulemans2020theoretical}, and models trained with standard backpropagation (BP), across three audiovisual benchmarks: AVE \cite{tian2018}, Kinetics-Sounds \cite{arandjelovic2017}, and VGGSound \cite{chen2020}. To isolate the effects of the respective learning rule, we deliberately fix early sensory processing and evaluate downstream associative plasticity on pretrained embeddings rather than using raw pixel values or waveforms. This ensures that any observed differences can be attributed to the learning rule itself and not to biases in upstream feature extraction. We then quantify representational efficiency in both unimodal and bimodal conditions by measuring mutual information, representational compression, and task-relevant accuracy. Our goal is to assess whether bimodal integration enables a more efficient allocation of representational resources compared to unimodal processing.

In summary, our study makes three contributions to computational neuroscience. First, we introduce an operational framework for comparing learning rules in terms of representational cost under biologically motivated structural constraints. Second, we test whether a strictly local excitatory Hebbian rule can generate compact, task-relevant representations without relying on global error transport. Third, we quantify the resulting trade-off between task-relevant information and retained input information, using CTI as a comparative proxy for representational resource allocation rather than as a direct measure of biological energy use. Our results indicate that Hebbian learning generates sparse, task-relevant representations that maintain high information content while reducing representational cost under a biologically motivated proxy, with these effects preserved under enforced sparsification. Importantly, we demonstrate that such local synaptic resource allocation can be achieved purely through well-established Hebbian updates, providing mechanistic insight into potential principles of neuronal resource assignment grounded in biologically plausible learning rules.

\section{Theoretical Background}
\label{fep}

\subsection{Variational Free Energy as a Biologically Grounded Objective}

\noindent Energetic constraints impose hard limits on the representational capacity of biological systems \cite{knoblauch, knoblauch2017}. Neural circuits operate within a subset of theoretically possible states, yet encode behaviorally and cognitively relevant information \cite{hasenstaub}. This constrains them to configurations consistent with likely sensory inputs, reducing uncertainty and promoting metabolically efficient representations.

These constraints can be formalized by the FEP \cite{friston2010free,fristonfree}, which describes how internal states approximate a posterior over environmental causes while minimizing sensory surprise, i.e., the negative log evidence $-\log p(x)$ of sensory inputs under a generative model. Since direct computation of $p(x)$ is generally intractable, the FEP introduces variational free energy as a tractable upper bound:
\begin{equation}
\mathcal{F}\bigl[q(z\mid x)\bigr]
\;=\;
\mathbb{E}_{q(z\mid x)}\!\bigl[-\log p(x,z)\bigr]
\;-\;
H\!\bigl[q(z\mid x)\bigr],
\label{fepfunctional}
\end{equation}
\noindent where $q(z\mid x)$ is an approximate posterior over latent causes $z$, and $H[q(z\mid x)]$ denotes the entropy of the approximate posterior. Equation~\eqref{fepfunctional} has the standard ``energy minus entropy'' form, in which the expected negative log-joint term penalizes unlikely latent explanations, whereas the entropy term favors broader, less committed posteriors.

For a generative model that factorizes as
\[
p(x,z)=p(x\mid z)\,p(z),
\]
\noindent the same variational free energy admits the standard decomposition
\begin{equation}
\mathcal{F}\bigl[q(z\mid x)\bigr]
=
\mathrm{KL}\!\bigl[q(z\mid x)\,\|\,p(z)\bigr]
\;-\;
\mathbb{E}_{q(z\mid x)}\bigl[\log p(x\mid z)\bigr],
\label{eq:varF_kl}
\end{equation}
\noindent which implies $\mathcal{F}\bigl[q(z\mid x)\bigr] \ge -\log p(x)$. The KL term quantifies representational complexity relative to the prior, whereas the expected log-likelihood term quantifies predictive accuracy. Minimizing $\mathcal{F}\bigl[q(z\mid x)\bigr]$ therefore implements a formal trade-off between representational \noindent efficiency and predictive accuracy and corresponds to approximate Bayesian inference when optimized with respect to $q$ \cite{fristonfree}.

Maintaining organized low-entropy configurations requires physical work. In classical thermodynamics, Helmholtz energy quantifies this work for a system at constant temperature \cite{helmholtz1882}. As established in Section \ref{introduction}, this thermodynamic formalism provides a theoretical analogy to neuronal constraints. Following \cite{sengupta}, the entropy term \(H[q(z\mid x)]\) in the variational free energy functional can be formally related to the diversity of accessible internal states, providing a link between representational complexity and minimal baseline metabolic cost. Minimizing \(\mathcal{F}[q(z\mid x)]\) thus limits the entropy of accessible internal states, thereby constraining baseline metabolic expenditure, while the predictive term preserves task-relevant accuracy.

\subsection{The Variational Information Bottleneck Theory}

\noindent Following \cite{fristonfree}, the FEP functional \eqref{eq:varF_kl} defines a normative objective for internal states that balances predictive accuracy with representational efficiency. While the FEP admits a thermodynamic interpretation of self-organizing systems, an equivalent formulation of the same principle arises in information theory. The IB framework \cite{tishby1999} formalizes this balance as a trade-off between compression and relevance. Learning is cast as the search for an internal code \(Z\) that compresses sensory input \(X\) while retaining information relevant to behaviorally significant goals \(Y\),
\begin{equation}
\mathcal{L}_{\mathrm{IB}}[q(z\mid x)] = \beta\,I(Z; X) - I(Z; Y),
\end{equation}
\noindent where \(I(\cdot;\cdot)\) denotes mutual information and the Lagrange multiplier \(\beta>0\) regulates the trade-off between compression \(I(Z;X)\) and relevance \(I(Z;Y)\). Minimizing \(\mathcal{L}_{\mathrm{IB}}\) therefore reduces representational complexity while promoting task-relevant information.

Under standard modeling assumptions, namely stationary data distributions, approximately Gaussian latent posteriors, and identification of the FEP recognition density with the IB encoder, minimization of variational free energy can be shown to be formally related to optimization of the IB objective \cite{fristonfree,tishby1999}.

This formal relation provides a theoretical basis for using VIB \cite{alemi} as a tractable variational approximation to the IB objective, allowing a principled and implementable quantification of FEP-implied representational trade-offs under the stated assumptions. The VIB implements the IB by replacing mutual-information terms with variational bounds and optimizing parametrized stochastic encoders and decoders to extract compressed latent codes relevant to the task. Compression is quantified through the expected Kullback–Leibler divergence to a chosen prior \(r(z)\):
\begin{equation}\label{eq:kl_theory}
\mathbb{E}_x \mathrm{KL}\big(q(z\mid x)\,\|\,r(z)\big) = I(Z;X) + \mathrm{KL}\big(q(z)\,\|\,r(z)\big),
\end{equation}
\noindent where \(q(z)=\mathbb{E}_x q(z\mid x)\) denotes the aggregated posterior. The expected KL therefore approximates the mutual information up to the divergence between the aggregated posterior and the prior. Together, these formal relations establish the VIB as an operational tool for quantifying representational compression and task relevance in computational models. While the VIB formalism does not compute energy use explicitly, it provides a principled, quantitative proxy linking compact, low-entropy latent representations to a theoretical reduction in metabolic cost under the stated assumptions.

\section{Materials and Methods}
\label{method}

\subsection{The Hebbian rule}
\label{rule}

\noindent The core Hebbian update is based on Oja's neural PCA rule \cite{Oja1989}, an extension of classical Hebbian plasticity \cite{hebb}. The rule studied here, originally described in \cite{neucomp}, is an adaptation of this framework with nonnegativity and additional normalization (see Methods, Sec.~\ref{architecture}). These constraints, together with the competitive dynamics inherent in neural PCA, produce sparse weight matrices that limit the number of functional synapses \cite{ijcnn}.

The synaptic weight update in the hidden layers generalizes Oja's single-neuron PCA rule to a network with multiple postsynaptic units \cite{oja1991}:
\begin{equation}
    \Delta w_{ij} = \eta z_j \bigl(x_i - \sum_{k} z_k w_{ik}\bigr),
\label{local}
\end{equation}
\noindent where \(x_i\) denotes the presynaptic input, \(\eta\) the learning rate, and 
\[
z_j = \phi\Bigl(\sum_u x_u w_{uj}\Bigr)
\]
\noindent is the postsynaptic activation. In the present implementation, \(\phi(\cdot)\) denotes the min--max rescaling used to enforce bounded nonnegative hidden activations. For the preactivation
\[
a_j = \sum_u x_u w_{uj},
\]
\noindent the activation function is defined as
\[
\phi(a_j)
=
\frac{a_j-\min_m a_m}{\max_m a_m-\min_m a_m+\epsilon},
\]
\noindent where the minimum and maximum are computed over the preactivations of the corresponding hidden layer, and \(\epsilon>0\) is a small numerical constant. Thus, \(z_j\in[0,1]\), ensuring bounded nonnegative activity for the subsequent Hebbian update. The subtractive competition term can be written as
\[
-\eta z_j \sum_k z_k w_{ik} = -\eta z_j (\mathbf{z}\cdot\mathbf{w}_i),
\]
\noindent where \(\mathbf{z}=(z_1,\dots,z_M)^\top\) denotes postsynaptic population activity and \(\mathbf{w}_i=(w_{i1},\dots,w_{iM})^\top\) are the outgoing synapses of the presynaptic neuron \(i\). Thus, the update depends on the signals available to the presynaptic neuron: the local input \(x_i\), the postsynaptic response \(z_j\), and the presynaptic projection \(\mathbf{z}\cdot\mathbf{w}_i\).

\cite{friston} proposed a computational model for implementing such dynamics in biological circuits. Mechanistically, the update combines a synapse-specific Hebbian potentiation of \(w_{ij}\) when \(x_i\) and \(z_j\) coincide, and a centrally computed decay proportional to \(z_j(\mathbf{z}\cdot\mathbf{w}_i)\) applied across that presynaptic neuron's outgoing synapses \cite{Lopez1990}. The result is two interacting, antisymmetrical influences on connectivity, with local consolidation occurring at individual synapses and weakening computed at the cell body and broadcast uniformly to that neuron's presynaptic terminals. Because consolidation is evaluated per synapse while weakening is computed at the presynaptic neuron level, correlated presynaptic inputs are strengthened, whereas synapses projecting to multiple postsynaptic targets are decorrelated and weakened. Although \eqref{local} is algebraically symmetric, this separation of potentiation and decay breaks practical symmetry and produces differential synaptic outcomes. These correspondences motivate the learning rule as a functional analogy to cortical plasticity without implying a direct mechanistic identity.

\subsection{Variational Information Bottleneck Implementation}
\label{encoder}

\noindent Throughout the theoretical discussion in Section~\ref{fep}, $x$ and the corresponding random variable $X$ denote sensory data in the FEP sense. In our experiments, however, we do not apply the stochastic bottleneck to raw sensory streams. Instead, the VIB module is trained post hoc on the frozen last-hidden representation $h=f_\theta(x)$ of a trained deterministic MLP. Accordingly, in all information-based metrics we operationalize the IB input variable as the random variable $H$ induced by $h$ over the dataset, and we report compression using $I(Z;H)$, estimated via the standard VIB bound $\mathbb{E}_{h}\,\mathrm{KL}\!\left[q(z\mid h)\,\|\,r(z)\right]$.

Concretely, after training under each evaluated learning rule, the final classification layer is removed and replaced by a stochastic latent layer $Z$ that takes as input the activations $h$ of the last hidden layer of the pretrained MLP \cite{b22}. All preceding MLP weights remain fixed during VIB optimization. Only the parameters of the stochastic encoder and of the decoder are updated \cite{bwhpc}. The compression–relevance trade-off is controlled by the Lagrange multiplier $\beta>0$, as in the standard IB formulation.

The stochastic latent layer \(Z\) serves as a variational encoder by mapping the deterministic activations \(h\) of the last hidden layer \(H\) to a diagonal Gaussian
\[
q(z\mid h)=\mathcal{N}\!\big(\mu(h),\operatorname{diag}(\sigma(h)^2)\big),
\]
\noindent with \(\mu(h)\) and an unconstrained scale \(\rho(h)\) produced by a single trainable linear layer that outputs \(2K\) values for latent dimension \(K\). Standard deviations are obtained as
\[
\sigma(h)=\operatorname{softplus}\big(\rho(h)-5.0\big),
\]
\noindent where the offset of $5.0$ is subtracted to ensure small initial standard deviations, stabilizing the stochastic latent layer \cite{alemi}. Sampling uses the reparameterization trick,
\[
z=\mu(h)+\sigma(h)\odot\epsilon,\qquad \epsilon\sim\mathcal{N}(0,I),
\]
\noindent which yields unbiased gradient estimates for the encoder and decoder parameters \cite{kingma2014}.

The decoder \(q(y\mid z)\) is implemented as a linear classifier that maps sampled \(z\) to class probabilities and is used solely for evaluation of predictive performance. The VIB loss applied to the frozen representations is
\begin{equation}\label{eq:vib_loss_methods_final}
\mathcal{L}_{\mathrm{VIB}} = \frac{1}{N} \sum_{n=1}^{N} \mathbb{E}_{z \sim q(z \mid h_n)}[-\log q(y_n \mid z)] + \beta\, \mathbb{E}_h \mathrm{KL}\big(q(z \mid h) \,\|\, r(z)\big),
\end{equation}
\noindent with \(r(z) = \mathcal{N}(0,I)\) unless otherwise stated. The KL term is computed as in Eq.~\eqref{eq:kl_theory}, serving as a tractable proxy for \(I(Z;H)\) up to the divergence between the aggregated posterior and the prior. Importantly, this loss is not used to train the underlying MLP with Hebbian or other biologically plausible rules. It is applied post hoc only to train the stochastic VIB encoder and decoder on the frozen representations, in order to quantify task-relevant information and representational cost.

In practice, the expectation over \(z\) in Eq.~\eqref{eq:vib_loss_methods_final} is approximated via Monte Carlo sampling with \(S=12\) draws per input \(h_n\). For evaluation, \(I(Z;H)\) was approximated by the mean KL divergence to the prior, whereas \(I(Z;Y)\) was estimated using the variational lower bound \(H(Y)-\mathbb{E}_{q(z\mid h)}[-\log q(y\mid z)]\), with the expectation approximated using the same Monte Carlo samples. During VIB optimization, only the encoder and decoder parameters are updated, while the pretrained MLP weights remain frozen. This implementation follows the VIB formulation of \cite{alemi} and the reparameterization procedure of \cite{kingma2014}, adapted to operate on the post hoc frozen representations of the pretrained MLP. Only the VIB component, comprising the stochastic encoder and its training procedure, is directly based on the reference implementation provided by \cite{alemi}\footnote{\url{https://github.com/alexalemi/vib_demo}}, ensuring accurate estimation of mutual information and reproducibility. All other components, procedures, and analyses were developed independently. Hyperparameters are provided in Section~\ref{architecture}.

\subsection{Datasets and Preprocessing}
\label{datasets}

\noindent We evaluate the Hebbian learning rule on three widely used audiovisual benchmarks: AVE \cite{tian2018}, Kinetics-Sounds \cite{arandjelovic2017}, and VGGSound \cite{chen2020}. These datasets provide a standardized basis for assessing multimodal representation learning given the limited availability of large-scale annotated audiovisual corpora. 

The AVE dataset contains 4,143 video clips of ten seconds each across 28 audio-visual event classes. Kinetics-Sounds, a 30-class subset of Kinetics-400 \cite{kay2017}, comprises approximately 24,000 ten-second videos. For VGGSound \cite{chen2020}, we use the VGGSound100 split defined by \cite{pian2023}, consisting of 60,000 video clips from 100 randomly selected classes with balanced validation and test sets.

We adopt the feature-extraction pipeline of \cite{pian2023}. The pipeline, the pre-extracted visual and audio features, and the data splits are available at the project repository\footnote{\url{https://github.com/weiguoPian/AV-CIL_ICCV2023}}. Each video clip is represented, for each modality, as a temporal sequence of embeddings of dimension $D$, forming a matrix $E \in \mathbb{R}^{T\times D}$. The embeddings are generated by pretrained VideoMAE and AudioMAE models \cite{tong2022, huang2022}. For bimodal inputs, the audio and visual embeddings are concatenated along the feature dimension at each time step, resulting in $E \in \mathbb{R}^{T\times (768+768)}$.

All experiments operate on these pretrained embeddings rather than on raw pixels or waveforms. We interpret them as high-level, task-relevant representations, consistent with prior work demonstrating that task-optimized deep network features can predict higher-level cortical responses in vision and audition \cite{Yamins2014,Kell2018}.  Within this framework, the audio and visual embeddings are treated as high-level, task-relevant features produced by a fixed upstream encoder \cite{he2022}. This approximates a setting in which early sensory processing is held constant while downstream associative learning is varied. We do not claim a one-to-one correspondence between MAE representations and specific cortical areas. Nevertheless, this abstraction allows analysis of synaptic resource allocation across modalities under anatomically and metabolically inspired constraints, while avoiding confounds introduced by the architectural inductive biases required for raw-input feature extraction.

To obtain fixed-length clip-level representations for downstream fully connected classifiers, we aggregate across time by computing the mean embedding
\begin{equation}
\bar{e}
=
\frac{1}{T}\sum_{t=1}^{T} e_t
\in
\mathbb{R}^{D},
\label{eq:clip_mean}
\end{equation}
\noindent where $e_t \in \mathbb{R}^{D}$ denotes the embedding at time step $t$ and $D=768$ for unimodal inputs or $D=1536$ for bimodal inputs. We denote the resulting fixed-length clip-level input by $x := \bar{e}$.

Although averaging reduces temporal resolution, preliminary tests demonstrated that the clip-level input $x$ still enables reliable audiovisual classification. This design emphasizes our main goal of assessing the efficiency and structure of learned representations, including information content, sparsity, and compressibility, rather than benchmarking absolute predictive performance. Conceptually, this temporal aggregation can be viewed as analogous to cortical pooling mechanisms in higher-order areas, which integrate neural activity over extended timescales to form invariant representations \cite{hasson2008hierarchy, honey2012}.

To ensure stable network training, the clip-level input $x$ is preprocessed before being fed into the models. For nonnegativity-constrained networks, namely Hebbian and constrained BP, min–max normalization is applied. For networks without such constraints, i.e. standard BP and DDTP, rescaling the inputs as
\[
x_\text{scaled} = \frac{x - 0.5}{0.5}
\]
\noindent was necessary to stabilize training, particularly in deep architectures, where standard min–max normalization led to poor convergence or nonconvergent behavior. This adjustment ensures reliable training across all network types while maintaining compatibility with their respective activation functions.

\subsection{Sparsification}
\label{sparse}

\noindent Sparse neural networks can be obtained via two primary approaches, dense-to-sparse and sparse-to-sparse. In the dense-to-sparse paradigm, a fully dense network is trained initially, allowing the model to exploit its full representational capacity, after which redundant or less important connections are pruned according to criteria such as weight magnitude, sensitivity, or regularization \cite{atashgahi2022}. Methods include pruning after training \cite{han2015learning}, pruning during training via regularization techniques such as $L_0$ penalties \cite{louizos2018learning}, and pruning before training based on connection sensitivity or flow metrics \cite{tanaka2020pruning}. These approaches generally yield sparse networks that retain accuracy comparable to dense networks. By contrast, sparse-to-sparse training initializes the network with sparse connectivity, either static \cite{dettmers2019sparse} or dynamically evolving \cite{evci2020rigl}. While computationally efficient, sparse-to-sparse methods require careful tuning of connectivity dynamics and may underutilize the initial representational capacity. 

Some learning rules, particularly those constrained by nonnegativity and competition, including those evaluated in this study, yield inherently sparse synaptic connectivity \cite{ijcnn}. In contrast, gradient-based optimization typically does not induce sparsity without additional regularization. To ensure comparability across learning rules, we therefore impose the same sparsity structure on all networks via post hoc pruning.

From a biological perspective, dense-to-sparse training is appealing, as cortical circuits are initially highly overconnected and undergo activity-dependent synaptic pruning, removing redundant connections while preserving essential pathways for efficient processing \cite{braitenberg1998cortex, chechik1998synaptic}. This analogy suggests that dense training followed by pruning may better reflect natural learning processes and yield sparse representations that retain task performance under synaptic removal. In this study, we therefore adopt a dense-to-sparse strategy motivated by both computational efficiency and biological plausibility. Networks are trained for 500 epochs with their respective learning rules. Subsequently, layer-wise magnitude pruning is applied, where necessary, to remove the least important weights, targeting approximately 90\% sparsity, consistent with cortical observations \cite{braitenberg1998cortex, chechik1998synaptic}.

\subsection{Implementation Details and Training Protocol}
\label{architecture}

\noindent Learning rules were evaluated using fully connected MLPs in two architectural variants, shallow networks comprising a single hidden layer and deep networks with five hidden layers. The width of each hidden layer was set to match the input dimensionality, 768 units for unimodal inputs (audio or visual) and 1,536 units for bimodal inputs, obtained by concatenating VideoMAE and AudioMAE features. The output dimensionality was dataset dependent, with 28 classes for AVE, 30 classes for Kinetics-Sounds, and 100 classes for VGGSound100. Preliminary experiments confirmed that the qualitative behavior of the evaluated learning rules was consistent across a range of hidden layer widths and depths. The configurations reported here were selected to match the input dimensionality and ensure stable training.

To ensure stable propagation through multiple layers in networks with nonnegativity constraints, namely Hebbian learning or constrained BP, we apply a magnitude-preserving Z-score normalization to the input of each hidden layer. For a layer input \(x\), the normalized input \(\tilde{x}\) is defined as
\begin{equation}
\tilde{x}
=
\gamma
\frac{x-\mu_x}{\sigma_x+\epsilon},
\label{normalize}
\end{equation}
\noindent where \(\mu_x\) and \(\sigma_x\) denote the mean and standard deviation of the layer input computed along each feature dimension, \(\epsilon>0\) is a small numerical constant, and \(\gamma\) is a dynamically computed scaling factor that preserves the input magnitude \cite{b2}. The normalized input \(\tilde{x}\) is then propagated through the layer, and the resulting layer output is subsequently rescaled to the interval \([0,1]\) to enforce nonnegativity \cite{neucomp}. This procedure simultaneously zero-centers the layer input, ensuring that PCA-like Hebbian updates converge correctly \cite{oja1982simplified}, and enforces bounded correlation-based updates, preventing divergence or exploding signals in deep networks. Zero-centering is particularly critical for multilayer correlation-based learning, as Oja's rule theoretically converges to principal components only when inputs are mean-centered. Networks without nonnegativity constraints, such as unconstrained BP and DDTP, propagate inputs directly through the layer, using standard \(\tanh\) activations without normalization, unless otherwise stated.

All experiments were implemented using TensorFlow \cite{tensorflow2015-whitepaper}. Training used class-balanced batches: 1,000 samples for nonnegativity-constrained networks (Hebbian and BP with nonnegativity constraint) to stabilize mean activity estimates in the Hebbian-PCA rule, and twice the class size for unconstrained networks (standard BP and DDTP), which benefit from smaller batches scaled to class diversity to maintain stable gradients or target signals. Learning rates were set to $\eta = 1\times 10^{-3}$ for BP, $\eta = 5\times 10^{-5}$ for Hebbian networks, and $\eta = 1\times 10^{-2}$ for DDTP networks, with cosine decay applied over 500 epochs. BP and DDTP networks were optimized using the Adam optimizer \cite{kingma2014adam}. No dropout or other explicit regularization was applied unless otherwise stated. Hyperparameters, including learning rates and batch sizes, were initially explored by a brief randomized search on a validation subset, informing the reported configuration.

After training the deterministic hidden layers, a VIB module \cite{alemi,b22} was appended on top of the frozen representation for evaluation. The stochastic latent layer was then trained on the frozen MLP features, with dimension $K=256$, consistent with \cite{alemi} and \cite{b22}. \cite{alemi} conducted extensive evaluations of multiple bottleneck sizes and identified $256$ as providing an optimal trade-off between compression and predictive performance across a variety of image classification tasks. Initial experiments on the trained MLP representations prior to appending the stochastic VIB layer confirmed these results. For the BP and DDTP networks, both full dense and sparsified variants were trained to ensure methodological consistency. Only sparsified variants, corresponding to the biologically constrained condition, are reported in the main text.

\subsubsection{Training Algorithm}

\noindent Our experiments follow a two-stage protocol with an explicit separation between representation learning and information-bottleneck evaluation.

For Hebbian and other nonnegativity-constrained networks, we implement the constraints as explicit operations during training. First, we apply the magnitude-preserving Z-score normalization in \eqref{normalize} to each hidden-layer input, computed per batch and per feature dimension, which zero-centers the presynaptic activity required for PCA-like convergence. Second, after the normalized preactivation is propagated through the layer, we enforce nonnegative activations by rescaling the resulting layer output to the interval \([0,1]\) per batch. Third, after each weight update computed from \eqref{local}, we project weights onto the nonnegative orthant via \(W \leftarrow \max(W,0)\). Pruning is applied by masking weights according to Section~\ref{sparse}.

\begin{figure}[H]
    \centering
    \begin{algorithmic}[1]
    \STATE \textbf{Phase 1: Train deterministic MLP encoder with local rule}
    \STATE Initialize all MLP weights to small, nonnegative random values
    \FOR{each epoch}
        \STATE Shuffle training samples
        \FOR{each class-balanced batch of size $N$}
            \FOR{each hidden layer $l$}
                \STATE Apply \eqref{normalize} to hidden-layer inputs
                \STATE Forward pass and rescale hidden activations to \([0,1]\) to enforce nonnegativity
                \STATE Save hidden activations for local updates
                \STATE Compute $\Delta w^{l}_{ij}$ using \eqref{local}
                \STATE Update weights
                \STATE Project weights to nonnegative values \(W \leftarrow \max(0, W)\)
            \ENDFOR
        \ENDFOR
    \ENDFOR
    \STATE Apply sparsification mask to each hidden layer to enforce target connectivity (see Section~\ref{sparse})
    \STATE Discard original classification head and freeze encoder representation \(H\)
    \STATE
    \STATE \textbf{Phase 2: VIB evaluation on frozen representations}
    \STATE Append VIB module \(H \rightarrow Z\) with \(K=256\) and new classifier \(Z \rightarrow \hat{Y}\)
    \FOR{each epoch}
        \FOR{each class-balanced batch of size $N$}
            \STATE Forward pass through frozen encoder to obtain \(H\)
            \STATE Sample \(Z\) from the VIB posterior and compute \(\hat{Y}\)
            \STATE Update only VIB and classifier parameters by minimizing the VIB objective
        \ENDFOR
    \ENDFOR
    \end{algorithmic}
    \caption{Two-stage training protocol. Phase 1 learns deterministic representations under normalization, nonnegativity, and pruning; Phase 2 evaluates frozen representations by fitting a VIB bottleneck and new classifier.}
    \label{fig:spgd}
\end{figure}

\noindent For reference models such as standard BP or DDTP, we employ the same two-stage protocol, including Phase 1 representation learning, layer-wise sparsification, and Phase 2 VIB evaluation, with the only modification being the use of the respective learning rule in place of the Hebbian update.

\subsection{Metrics}

\subsubsection{Task-Information Cost}
\label{metrics}

To quantify the efficiency of latent representations, we introduce the Task-Information Cost metric (CTI):
\begin{equation}
CTI = \frac{I(Z;H)}{I(Z;Y)},
\label{eq:cti}
\end{equation}
\noindent where $H$ denotes the frozen last-hidden representation $h$ that serves as input to the VIB encoder (Section~\ref{encoder}),
and $I(Z;Y)$ denotes task-relevant information about the labels.

The metric is biologically motivated by metabolic constraints in neural systems, but operationalized here as an information-theoretic comparison of retained input information relative to task-relevant information. Lower values of \(I(Z;H)\) for a given \(I(Z;Y)\) indicate a more compact representation under the stated assumptions. Accordingly, CTI should be interpreted as a comparative proxy of representational cost rather than as a direct estimate of biological energy expenditure.

In our deterministic-label classification setting, $I(Z;Y)$ is typically much smaller than $I(Z;H)$ because $Y$ is low-entropy and $Z$ is a stochastic function of a high-dimensional representation. Empirically, we observe $I(Z;H)\ge I(Z;Y)$ across all reported operating points, so CTI is $\ge 1$ in practice. Values approaching 1 indicate that relatively little input-dependent information is retained per unit of task-relevant information, whereas larger values indicate a greater retained-information burden and therefore a less compact latent code.

To theoretically illustrate this concept, we consider a neuron in the primary visual cortex (V1) responsible for encoding edge orientation. When responses are selective, \(I(Z;H)\) remains close to \(I(Z;Y)\), indicating minimal redundancy and efficient coding. If neuronal responses are indiscriminate, \(I(Z;H)\) increases relative to \(I(Z;Y)\), corresponding to a greater number of distinct stored patterns, such as individual pixels treated as separate patterns. Under this proxy, maintaining these additional functional synaptic states corresponds to a greater representational burden. Consequently, CTI provides a comparative information-theoretic measure of coding efficiency, with values that can be related to biologically plausible metabolic constraints.

\paragraph{Statistical Analysis}
\label{statistical_test}
\noindent To quantify the reliability of CTI differences across matched experimental runs, we used paired nonparametric analyses across learning rules. Because CTI is a positive ratio-scale quantity, pairwise comparisons were performed on log-transformed CTI ratios rather than on raw CTI differences.

For each reference method \(m\), we computed
\begin{equation}
d_{c,s}^{(m)}
=
\log\!\left(
\frac{
CTI^{(m)}(c,s)
}{
CTI^{(\mathrm{Hebbian})}(c,s)
}
\right),
\end{equation}
\noindent where \(m\) indexes the non-Hebbian reference method, \(c\) denotes the matched experimental condition, and \(s\) indexes the random seed. Positive values indicate higher CTI for the reference method than for the Hebbian model.

The pairwise comparisons were evaluated using two-sided Wilcoxon signed-rank tests on the log-transformed CTI ratios \(d_{c,s}^{(m)}\) \cite{wilcoxon1945individual}. The resulting \(p\)-values were corrected across the two comparisons using the Holm procedure \cite{holm1979simple}. Effect sizes were reported as geometric CTI ratios,
\begin{equation}
R^{(m)}
=
\exp\!\left(
\frac{1}{N}
\sum_{c,s}
d_{c,s}^{(m)}
\right),
\end{equation}
\noindent where \(N\) denotes the number of matched condition--seed pairs. Undefined CTI values and single-run control configurations were excluded from inferential comparisons.

\subsubsection{Top-1 Accuracy}

\noindent Top-1 accuracy measures the fraction of samples for which the predicted class matches the ground truth. Importantly, all reported Top-1 accuracy values are obtained exclusively during Phase 2 of the training protocol (Fig.~\ref{fig:spgd}), i.e., after the deterministic encoder has been trained, frozen, and evaluated through the appended VIB module and its associated classifier. No performance metrics are computed from the Phase 1 classifier.

Its role here is primarily to ensure that sparse, metabolically constrained networks retain sufficient task-relevant functionality and do not achieve low CTI at the expense of rendering the learned representation ineffective.

In the present study, Top-1 accuracy is therefore treated as a feasibility constraint on the VIB-based evaluation model. Reductions in CTI are only meaningful if task performance remains non-trivial under the same VIB evaluation setting. Accordingly, we report Top-1 accuracy alongside CTI and emphasize comparisons of representational efficiency at comparable Phase 2 performance.

\section{Results}

\noindent This section presents a controlled mechanistic comparison of learning rules under matched architectural, sparsity, and evaluation constraints. The aim is not to identify the strongest audiovisual classifier, but to test whether a strictly local plasticity rule alters the trade-off between functional performance and representational cost.

To operationalize this investigation, we employ a two-stage protocol. In the first stage, MLPs are trained according to the respective learning rule under evaluation. In the second, post hoc stage, a VIB module is appended on top of the frozen representations. This stage quantifies how much predictive performance can be retained when the learned features are subjected to an explicit information bottleneck, independently of the original MLP training. Consequently, Top-1 accuracy is constrained by the information available in the frozen representation. We interpret each model's outcome as a point in the trade-off space between predictive performance, quantified by Top-1 accuracy, and representational cost, quantified by CTI. Here, Top-1 accuracy provides a measure of functional feasibility, whereas CTI quantifies retained input information normalized by task-relevant information, acting as a proxy for representational cost. The key question is whether a strictly local plasticity rule can shift this cost–performance trade-off relative to reference models under matched architectural and sparsity constraints.

To this end, we benchmark the Hebbian rule against two reference models: (1) DDTP \cite{meulemans2020theoretical}, which implements local target-based multilayer credit assignment without requiring symmetric weight transport. DDTP is selected because it (i) provides a biologically motivated, layer-wise learning rule, (ii) scales to moderately deep feedforward architectures and delivers competitive performance among biologically inspired algorithms, and (iii) features local update rules that are compatible with sparsification and employs a reconstruction loss to improve alignment of forward weight updates with layer-wise targets, making it directly comparable to the constrained, scalable training protocol studied here \cite{gupta2022}; and (2) a reference model trained with standard BP on the identical MLP encoder and stochastic latent layer described in Section~\ref{encoder}, providing an unconstrained upper bound on predictive performance.

In order to contextualize these benchmarks within the broader landscape of Hebbian learning, including Oja's normalized Hebbian rule and related subspace-learning variants, several extensions and related approaches have been reported. These include unsupervised feedforward models combining Hebbian synaptic and structural plasticity \cite{naresh}, as well as supervised Hebbian learning in deep counterstream associative networks \cite{knoblauch2026counterstream}. Classical two-stage schemes used generalized Hebbian rules to initialize internal representations before subsequent supervised training \cite{Karayiannis1996}. More recent hybrid approaches combine Oja's rule with BP or related error-driven updates to stabilize online learning, preserve activation norms, and maintain richer activation subspaces in deeper feedforward or recurrent networks \cite{ShervaniTabar2025}. These studies show that Oja-related learning can support multilayer supervised optimization, either as an unsupervised initialization procedure or as an auxiliary term in an error-driven update. The present work addresses a different question. We do not use the Hebbian rule as a pretraining initializer for later gradient-based fine-tuning, nor do we combine it with BP as a hybrid supervised update. Instead, we evaluate a constrained local Hebbian rule as the primary representation-learning mechanism and then apply a post hoc VIB readout to quantify task-relevant information and representational cost.

Other extensions include deep Hebbian architectures incorporating convolutional layers, such as FastHebb \cite{fasthebb}, classical and Winner-Takes-All (WTA) Hebbian rules for shallow networks \cite{hebb, hopfield19}, and modern self-supervised, contrastive, and self-distillation approaches \cite{chen2020simple, he2020momentum, Dong_2023_CVPR, Jang_2023_WACV}. Together, these approaches provide relevant context, but they do not provide controlled baselines for isolating the mechanism studied here because they either use Oja-related learning as initialization, combine it with supervised error-driven updates, rely on convolutional or self-supervised architectural priors, or omit explicit representational-cost analysis under the same VIB readout. We therefore use a controlled MLP assay and include direct GHA, classical Oja subspace, PCA-to-readout, and normalization/rescaling controls below.

To address these limitations, we employ a clearly defined MLP backbone as an assay design. This intentionally simplified setting isolates the effects of the learning rule on synaptic resource allocation while minimizing confounds from additional architectural priors. It enables a systematic mechanistic investigation of synaptic resource allocation and network dynamics, independent of architectural complexity. For rigorous assessment, all models are evaluated under approximately 10\% synaptic connectivity, unless stated otherwise, consistent with estimates of cortical connectivity \cite{attwell2001energy, harris2012synaptic}. We additionally report a dense BP model as an unconstrained reference upper bound on raw predictive performance.

\subsection{Quantitative Results}
\label{results}

\begin{figure*}[t] 
    \centering
    \includegraphics[width=\textwidth]{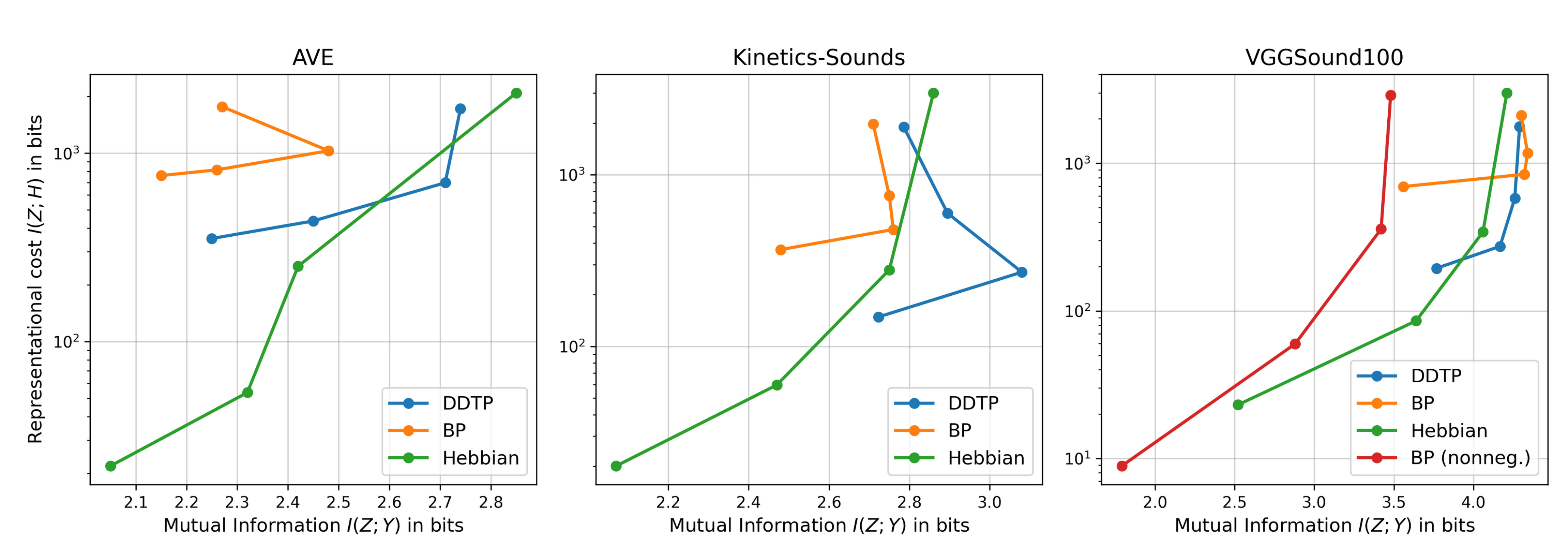} 
    \caption{Task-relevant information \(I(Z;Y)\) versus representational cost \(I(Z;H)\) for networks trained with DDTP, BP, and the Hebbian rule under study. Points are connected within each learning-rule trajectory according to increasing bottleneck strength. For each trajectory, the points are ordered from top to bottom as \(\beta=0,10^{-3},10^{-2},10^{-1}\), where \(\beta=0\) denotes the setting without a VIB compression penalty. The y-axis is log-scaled.}
    \label{fig:deep}
\end{figure*}

\subsubsection{Information Compression}

\noindent Fig.~\ref{fig:deep} summarizes the trade-off between task-relevant information \(I(Z;Y)\) and representational cost for deep networks. Specifically, we plot \(I(Z;Y)\) against a proxy for representational cost given by \(I(Z;H)\), the mutual information between \(Z\) and the frozen MLP representation \(H\). The quantities are estimated using the KL-based proxy for \(I(Z;H)\) and the variational lower bound for \(I(Z;Y)\) described in Section~\ref{encoder}. Results are shown for audiovisual inputs on AVE, Kinetics-Sounds, and VGGSound100. We show only deep configurations because they exhibit the clearest separation in compression behavior across learning rules. Shallow networks follow the same qualitative trends and are omitted for clarity. Nonnegativity-constrained BP is included only for VGGSound100.

Across datasets and \(\beta\) values, BP and DDTP generally achieve higher \(I(Z;Y)\), but this performance is accompanied by higher \(I(Z;H)\), indicating that their predictive information is supported by representations that retain more input-dependent information. In contrast, Hebbian-trained representations typically attain comparable levels of \(I(Z;Y)\) at lower \(I(Z;H)\), implying a more favorable information-efficiency trade-off under the same architectural and sparsity constraints. Finally, AVE and Kinetics-Sounds exhibit higher absolute values of \(I(Z;Y)\) across methods, consistent with these tasks being easier relative to the capacity of the deep architectures, which reduces the pressure to discard task-relevant information even under stronger bottleneck regularization.

Interestingly, $I(Z;Y)$ initially increases at moderate bottleneck values ($\beta$) before declining under stronger compression. This transient increase indicates that networks can discard redundant or noisy information, consistent with prior observations that moderate bottlenecks improve generalization \cite{alemi}. Further increases in $\beta$, corresponding to stronger bottlenecks, lead to a rapid decline in performance for DDTP and BP, and these operating points were consequently excluded from presentation.

Table~\ref{res1} complements Fig.~\ref{fig:deep} by reporting Top-1 accuracy after the Phase-2 VIB readout together with the corresponding task-information cost CTI across datasets, architectures, and learning rules for the intermediate bottleneck setting \(\beta=10^{-2}\). This setting was selected as the main tabular comparison because it represents the compressed regime of interest while preserving sufficient task-relevant information for the downstream readout. Dense BP generally attains strong predictive performance, but often at higher CTI, particularly in the deep architecture. Sparse BP and DDTP remain competitive in several settings, although their information costs vary considerably across datasets and network depth. In contrast, Hebbian-trained representations do not uniformly maximize Top-1 accuracy, but they consistently occupy a low-CTI regime across shallow and deep architectures. This indicates that the proposed constrained Hebbian learning rule primarily improves the representational-cost profile rather than providing a uniform accuracy advantage.

\begin{table}[H]
\begin{center}
\centering
\caption{Top-1 accuracy [\%] after the Phase-2 VIB readout and task-information cost (CTI, unitless) for the AVE, Kinetics-Sounds, and VGGSound100 datasets using shallow and deep architectures, reported for \(\beta = 10^{-2}\). Results are reported as mean \(\pm\) standard deviation over five independent seeds, except where indicated by \(\dagger\). The Hebbian method follows \cite{neucomp}, and DDTP follows \cite{meulemans2020theoretical}.}
\resizebox{\textwidth}{!}{%
\begin{tabular}{l c |cc|cc|cc}
\hline
& 

& \multicolumn{2}{c|}{\textbf{AVE}} 
& \multicolumn{2}{c|}{\textbf{Kinetics-Sounds}} 
& \multicolumn{2}{c}{\textbf{VGGSound100}} \\
\cline{3-8}
\textbf{Method} & \textbf{Arch} & Top-1 [\%] & CTI & Top-1 [\%] & CTI & Top-1 [\%] & CTI \\
\hline
Standard BP \textsuperscript{*} & Shallow & \(64.1 \pm 0.5\) & \(232 \pm 16\) & \(63.9 \pm 1.1\) & \(30 \pm 3\) & \(62.8 \pm 0.7\) & \(26 \pm 2\) \\
         & Deep & \(62.5 \pm 0.9\) & \(344 \pm 21\) & \(61.5 \pm 0.4\) & \(215 \pm 16\) & \(62.7 \pm 0.4\) & \(237 \pm 11\) \\
\hline
BP       & Shallow & \(65.7 \pm 1.0\) & \(214 \pm 7\) & \(54.3 \pm 0.6\) & \(19 \pm 1\) & \(62.8 \pm 0.6\) & \(25 \pm 2\) \\
         & Deep & \(54.2 \pm 1.6\) & \(344 \pm 25\) & \(62.0 \pm 0.7\) & \(202 \pm 10\) & \(59.1 \pm 0.7\) & \(189 \pm 9\) \\
\hline
DDTP & Shallow & \(66.5 \pm 1.0\) & \(210 \pm 8\) & \(68.9 \pm 1.0\) & \(172 \pm 7\) & \(62.7 \pm 0.5\) & \(100 \pm 6\) \\
         & Deep & \(52.9 \pm 0.8\) & \(161 \pm 10\) & \(63.8 \pm 0.6\) & \(89 \pm 5\) & \(57.2 \pm 0.4\) & \(101 \pm 10\) \\
\hline
BP (nonneg.) & Shallow & \(58.9 \pm 0.4\) & \(22 \pm 1\) & \(56.6 \pm 1.4\) & \(18 \pm 1\) & \(56.0 \pm 1.9\) & \(18 \pm 1\) \\
             & Deep & \(8.0 \pm 0.0\) & -- & \(5.0 \pm 0.0\) & -- & \(2.0 \pm 0.0\) & -- \\
\hline
Hebbian & Shallow & \(58.7 \pm 0.7\) & \(21 \pm 1\) & \(61.2 \pm 1.6\) & \(12 \pm 0\) & \(60.7 \pm 0.4\) & \(21 \pm 2\) \\
             & Deep & \(58.9 \pm 1.1\) & \(22 \pm 0\) & \(60.1 \pm 0.7\) & \(23 \pm 1\) & \(56.9 \pm 1.6\) & \(21 \pm 1\) \\
\hline
BP (nonneg.) & Deep\textsuperscript{*} & -- & -- & -- & -- & -- & -- \\
\textit{Skip Con.} & Deep & -- & -- & -- & -- & \(44.5^{\dagger}\) & \(21^{\dagger}\) \\
\hline
\end{tabular}%
}
\label{res1}
\footnotesize{\textsuperscript{*}Dense connectivity; all other methods use sparse connectivity. Missing entries indicate unavailable or undefined results. \(\dagger\)Only one valid run was available; no standard deviation is reported.}
\end{center}
\end{table}

\noindent Nonnegativity-constrained BP substantially reduces representational cost relative to standard BP in the shallow architecture and reaches the same low-CTI regime as the Hebbian rule. However, this reduction is accompanied by lower Top-1 accuracy on Kinetics-Sounds and VGGSound100, and the corresponding deep nonnegative-BP configurations yield chance-level performance with undefined CTI. Thus, nonnegativity can reduce retained information in shallow networks, but does not reproduce the stable accuracy--cost profile observed for the constrained Hebbian rule across architectures. Across the evaluated bottleneck settings, Hebbian-trained models therefore exhibit a more favorable trade-off between predictive information \(I(Z;Y)\) and compression of the encoder representation \(I(Z;H)\) than standard BP, sparse BP, and DDTP, while remaining comparable in representational cost to shallow nonnegativity-constrained BP.

A paired seed-level CTI analysis supported this pattern for the values underlying Table~\ref{res1}. Across matched dataset--architecture conditions with defined CTI values, sparse BP and DDTP showed higher CTI than the Hebbian model, with geometric CTI ratios of \(5.37\) and \(6.73\), respectively. Both contrasts remained significant after Holm correction using two-sided Wilcoxon signed-rank tests \((N=30, p_{\mathrm{Holm}}=3.73\times 10^{-9})\).

\begin{table}[H]
\begin{center}
\centering
\caption{Top-1 accuracy [\%] after the Phase-2 VIB readout and task-information cost (CTI, unitless) for visual (Vis), auditory (Aud), and bimodal (Bi) modalities on the VGGSound100 dataset using the deep MLP variant (five hidden layers), for \(\beta = 10^{-3}, 10^{-2}, 10^{-1}\). Results are reported as mean \(\pm\) standard deviation over five seeds, except where indicated by \(\dagger\). The Hebbian method follows \cite{neucomp}, and DDTP follows \cite{meulemans2020theoretical}.}
\resizebox{\textwidth}{!}{%
\begin{tabular}{l c |cc|cc|cc}
\hline
& 

& \multicolumn{2}{c|}{\textbf{Vis}} 
& \multicolumn{2}{c|}{\textbf{Aud}} 
& \multicolumn{2}{c}{\textbf{Bi}} \\
\cline{3-8}
\textbf{Method} & \textbf{Beta (\(\beta\))} & Top-1 [\%] & CTI & Top-1 [\%] & CTI & Top-1 [\%] & CTI \\
\hline
Standard BP \textsuperscript{*} & $1 \times 10^{-3}$ & \(37.5 \pm 0.4\) & \(531 \pm 2\) & \(46.5 \pm 0.3\) & \(351 \pm 15\) & \(61.9 \pm 0.3\) & \(308 \pm 7\) \\
         & $1 \times 10^{-2}$ & \(37.6 \pm 0.5\) & \(382 \pm 16\) & \(46.3 \pm 0.5\) & \(288 \pm 9\) & \(62.7 \pm 0.4\) & \(237 \pm 11\) \\
         & $1 \times 10^{-1}$ & \(35.6 \pm 0.5\) & \(433 \pm 37\) & \(44.0 \pm 0.6\) & \(312 \pm 16\) & \(62.4 \pm 0.4\) & \(248 \pm 8\) \\
\hline
BP       & $1 \times 10^{-3}$ & \(33.2 \pm 0.4\) & \(355 \pm 8\) & \(41.6 \pm 0.4\) & \(333 \pm 6\) & \textbf{\(58.5 \pm 0.5\)} & \(277 \pm 11\) \\
         & $1 \times 10^{-2}$ & \(33.9 \pm 0.4\) & \(206 \pm 5\) & \(42.6 \pm 0.3\) & \(227 \pm 8\) & \textbf{\(59.1 \pm 0.7\)} & \(189 \pm 9\) \\
         & $1 \times 10^{-1}$  & \(31.6 \pm 0.5\) & \(171 \pm 9\) & \(38.2 \pm 0.9\) & \(218 \pm 6\) & \textbf{\(55.7 \pm 0.8\)} & \(192 \pm 2\) \\
\hline
DDTP & $1 \times 10^{-3}$ & \(34.3 \pm 1.0\) & \(257 \pm 10\) & \(39.7 \pm 0.6\) & \(199 \pm 6\) & \(57.6 \pm 0.4\) & \(144 \pm 6\) \\
         & $1 \times 10^{-2}$  & \(34.5 \pm 0.3\) & \(95 \pm 5\) & \(37.3 \pm 0.3\) & \(80 \pm 7\) & \(57.2 \pm 0.4\) & \(101 \pm 10\) \\
         & $1 \times 10^{-1}$   & \(30.7 \pm 0.5\) & \(49 \pm 1\) & \(33.3 \pm 0.3\) & \(42 \pm 1\) & \(54.6 \pm 0.4\) & \(46 \pm 5\) \\
\hline
BP (nonneg.)\textsuperscript{\(\dagger\)} & $1 \times 10^{-3}$  & \(33.2\) & \(109\) & \(41.2\) & \(92\) & \(48.8\) & \(105\) \\
\textit{Skip Con.} & $1 \times 10^{-2}$ & \(31.6\) & \(22\) & \(39.4\) & \(18\) & \(44.5\) & \(21\) \\
             & $1 \times 10^{-1}$ & \(24.6\) & \(6\) & \(32.2\) & \(5\) & \(34.6\) & \textbf{\(5\)} \\
\hline
Hebbian & $1 \times 10^{-3}$ & \(31.1 \pm 0.5\) & \(188 \pm 4\) & \(35.8 \pm 1.1\) & \(115 \pm 4\) & \(57.7 \pm 1.0\) & \textbf{\(79 \pm 3\)} \\
             & $1 \times 10^{-2}$ & \(31.5 \pm 0.3\) & \(31 \pm 1\) & \(33.9 \pm 0.7\) & \(20 \pm 1\) & \(56.9 \pm 1.6\) & \textbf{\(22 \pm 1\)} \\
             & $1 \times 10^{-1}$ & \(29.9 \pm 0.2\) & \(16 \pm 1\) & \(30.9 \pm 0.3\) & \(9 \pm 1\) & \(51.1 \pm 0.2\) & \(9 \pm 0\) \\
\hline
\end{tabular}%
}
\label{res3}
\footnotesize{\textsuperscript{*}Dense connectivity; all other methods use sparse connectivity. \(\dagger\)Only one valid run was available; no standard deviation is reported.}
\end{center}
\end{table}

\noindent This pattern is most visible in configurations where Top-1 accuracy is similar across methods. In the shallow VGGSound100 setting, dense BP, sparse BP, and DDTP reach closely matched Top-1 accuracies of \(62.8 \pm 0.7\%\), \(62.8 \pm 0.6\%\), and \(62.7 \pm 0.5\%\), respectively, but differ substantially in CTI. Dense and sparse BP operate at \(26 \pm 2\) and \(25 \pm 2\), whereas DDTP requires \(100 \pm 6\). Hebbian-trained representations achieve \(60.7 \pm 0.4\%\) Top-1 accuracy in the same setting and remain in the low-cost regime with CTI of \(21 \pm 2\). These results indicate that several learning rules can support comparable readout performance, but differ markedly in the representational cost associated with that performance. The deep nonnegativity-constrained BP model shows strongly degraded Top-1 accuracy, indicating that this constraint combination is not stable in the corresponding deep setting.

Table~\ref{res3} provides the corresponding modality- and bottleneck-level analysis for the deep VGGSound100 setting. Across \(\beta\) values, bimodal readouts generally provide the strongest Top-1 accuracy, while the visual-only and auditory-only conditions identify the relative contributions of each sensory stream. The higher bimodal Top-1 accuracy is not accompanied by a uniform increase in CTI across the displayed configurations. In several method--\(\beta\) conditions, joint audiovisual readouts preserve task-relevant performance at lower or comparable CTI relative to unimodal alternatives. This indicates that the combined representation can exploit complementary audiovisual information without necessarily producing a proportional increase in representational cost.

Together, Fig.~\ref{fig:deep}, Table~\ref{res1}, and Table~\ref{res3} support a consistent interpretation. BP and DDTP can retain high task-relevant information, but this is often associated with higher representational cost. Hebbian-trained models tend to preserve task-relevant information at lower CTI, with the remaining differences in Top-1 accuracy depending on dataset, architecture, bottleneck strength, and input modality.

\subsubsection{Top-1 Classification Accuracy Across Architectures and Learning Rules}

\begin{table}[H]
\begin{center}
\centering
\caption{Top-1 accuracy [\%] after the Phase-2 VIB readout for visual (Vis), auditory (Aud), and bimodal (Bi) modalities across three datasets (AVE, Kinetics-Sounds, VGGSound100) and two neural architectures (shallow and deep) with dense or sparse connectivity, evaluated under each learning rule, reported for \(\beta = 0\). The Hebbian method follows \cite{neucomp}, and DDTP follows \cite{meulemans2020theoretical}.}
\resizebox{\textwidth}{!}{%
\begin{tabular}{l c|ccc|ccc|ccc}
\hline
& & \multicolumn{3}{c|}{\textbf{AVE Top-1 [\%]}} & \multicolumn{3}{c|}{\textbf{K-S Top-1 [\%]}} & \multicolumn{3}{c}{\textbf{VS100 Top-1 [\%]}} \\
\cline{3-11}
\textbf{Method} & \textbf{Arch} & Vis & Aud & Bi & Vis & Aud & Bi & Vis & Aud & Bi \\
\hline
Standard BP \textsuperscript{*} & Shallow & 46.6 & 47.3 & 65.6 & 54.2 & 44.1 & 65.1 & 39.3 & 45.1 & 61.7 \\
 & Deep & 46.9 & 47.2 & 61.7 & 49.8 & 41.3 & 61.2 & 36.8 & 46.0 & 62.2 \\
\hline
BP & Shallow & 49.0 & 47.5 & 63.9 & 52.3 & 42.9 & 59.5 & 38.6 & 43.3 & 61.9 \\
 & Deep & 35.9 & 36.8 & 53.3 & 47.4 & 41.9 & 61.3 & 33.4 & 42.4 & 57.9 \\
\hline
DDTP & Shallow & 46.6 & 45.1 & 62.8 & 55.8 & 44.2 & 68.1 & 40.7 & 45.2 & 61.7 \\
 & Deep & 40.5 & 38.0 & 52.9 & 49.8 & 38.7 & 63.2 & 34.3 & 38.1 & 57.2 \\
\hline
BP (nonneg.) & Shallow  & 45.9 & 47.6 & 59.2 & 50.7 & 38.4 & 63.3 & 33.8 & 34.7 & 58.1 \\
 & Deep  & -- & -- & 8.0 & -- & -- & 5.0 & -- & -- & 2.0 \\
 \hline
BP (nonneg.) & Deep\textsuperscript{*}  & -- & -- & -- & -- & -- & -- & 39.7 & 46.7 & 59.1 \\
\textit{Skip Con.} & Deep  & -- & -- & -- & -- & -- & -- & 33.0 & 41.2 & 48.9 \\
\hline
Hebbian & Shallow & 42.1 & 44.8 & 58.5 & 51.4 & 39.4 & 62.3 & 37.5 & 41.4 & 60.6 \\
 & Deep & 37.3 & 42.7 & 58.8 & 46.2 & 37.7 & 59.3 & 31.2 & 37.2 & 55.6  \\
\hline
\end{tabular}%
}
\label{res2}
\footnotesize{\textsuperscript{*}Dense connectivity; all other methods use sparse connectivity. Missing entries indicate unavailable or undefined results.}
\end{center}
\end{table}

\noindent Table~\ref{res2} reports Top-1 classification accuracy after the Phase-2 VIB readout for visual, auditory, and bimodal inputs across the evaluated datasets, architectures, and learning rules at \(\beta=0\). This condition removes the VIB compression penalty and is therefore used as an empirical predictive reference for the representational pipeline under the same architectural and sparsity constraints. These values are accordingly included as a functional feasibility check rather than as the primary optimization criterion. All learning rules, with the exception of standard BP,\footnote{We note that the performance of networks trained end-to-end with BP can be significantly enhanced using dropout, noise injection, data augmentation, and related standard techniques.} were sparsified as described in Section~\ref{sparse}.

We also evaluated the canonical DTP method \cite{lee2015difference}, its extended variant with Dynamic Routing \cite{meulemans2020theoretical}, and related feedback-alignment methods such as Direct Feedback Alignment (DFA) \cite{nokland2016direct}. These approaches consistently performed comparably to, or worse than, DDTP. Accordingly, only the best-performing DTP variant is reported. We additionally evaluated alternative Hebbian variants under strictly local constraints, including the WTA Hebbian variant proposed by \cite{hopfield19} and standard Hebbian learning. Consistent with prior observations \cite{neucomp}, these approaches do not scale reliably to deeper networks, preventing a direct comparison to scalable deep network methods. Consequently, they are not included in Table~\ref{res2}.

\paragraph{Top-1 Accuracy of Shallow Networks Across Learning Rules}

\noindent For shallow networks, mean Top-1 accuracy is dataset-dependent across all learning rules. On VGGSound100, the sparse baselines and Hebbian model yield comparable mean bimodal performance, with BP, DDTP, and Hebbian reaching \(61.9\%\), \(61.7\%\), and \(60.6\%\), respectively, whereas nonnegativity-constrained BP reaches \(58.1\%\). On AVE, the Hebbian model achieves lower mean bimodal accuracy than BP and DDTP (\(58.5\%\) vs. \(63.9\%\) and \(62.8\%\)), while remaining close to nonnegativity-constrained BP (\(59.2\%\)). On Kinetics-Sounds, Hebbian performance remains within the range of the constrained baselines (\(62.3\%\) vs.\ \(59.5\%\) for BP and \(63.3\%\) for nonnegativity-constrained BP), although DDTP performs substantially better in this setting \((68.1\%\)). Thus, in the constrained setting studied here, the shallow-network results do not indicate a uniform mean-accuracy advantage for any learning rule. Instead, they show that Hebbian representations remain task-informative across datasets while exhibiting dataset-specific trade-offs relative to sparse BP, DDTP, and nonnegativity-constrained BP. Notably, DDTP performs particularly well on Kinetics-Sounds, where it exceeds dense standard BP in mean shallow bimodal accuracy (\(68.1\%\) vs. \(65.1\%\)).

\paragraph{Impact of Increased Network Depth on Top-1 Accuracy}

\noindent Increasing network depth generally reduces Top-1 classification accuracy for the sparse and biologically constrained learning rules. Under the stochastic encoder-decoder architecture used here, deeper networks can accumulate information loss before the decoder readout, reducing the expressiveness of the final representation. This interpretation is consistent with prior work showing that task-relevant information may progressively decrease across depth before reaching the classifier \cite{silva2025, achille2018}. In the present experiments, the depth-related accuracy reduction is therefore interpreted as a feasibility constraint of the evaluated representation-learning setting rather than as the primary performance result.

It was not feasible to train standard BP with a nonnegativity constraint across all datasets using the same deep architecture as the Hebbian rule.\footnote{Training deep feedforward networks under strict nonnegativity constraints exhibited substantially more severe optimization and stability issues than the other reference methods considered here. In pilot experiments, the five-layer nonnegative architecture failed to converge on all datasets without a targeted stabilization protocol involving residual projection, skip connections, high-rate dropout, Gaussian feature noise, and careful hyperparameter tuning, with training yielding near-random performance across all datasets. Because the required tuning is dataset-specific and computationally prohibitive, reproducing stable deep nonnegative reference models for all datasets was deemed out of scope. Therefore, we report a single, conservatively tuned deep nonnegative BP reference model on VGGSound100, which is the most complex dataset and thus provides the most informative reference point for deep nonnegative performance.} As a consequence, the Top-1 accuracy of the deep nonnegative BP architecture remains near chance across the three datasets. The dense nonnegative variant with skip connections and additional stabilization achieves competitive unimodal and bimodal performance on VGGSound100, but its performance decreases substantially under sparsification. Overall, the Top-1 results indicate that nonnegativity constraints can impose a substantial optimization burden in deeper architectures, consistent with prior work showing that strictly nonnegative networks have a more restricted parameter space and can exhibit reduced function approximation capacity \cite{mikulincer2022, liu2020certified}. While nonnegativity supports interpretability, it can incur a reduction in classification performance \cite{ijcnn, b1}.

\subsubsection{Evaluation of Synaptic Allocation Efficiency}
\label{ablation_text}

\begin{figure*}[t]
    \centering
    \includegraphics[
        width=1.05\textwidth,
        trim=7.5cm 0cm 7.5cm 0cm, 
        clip
    ]{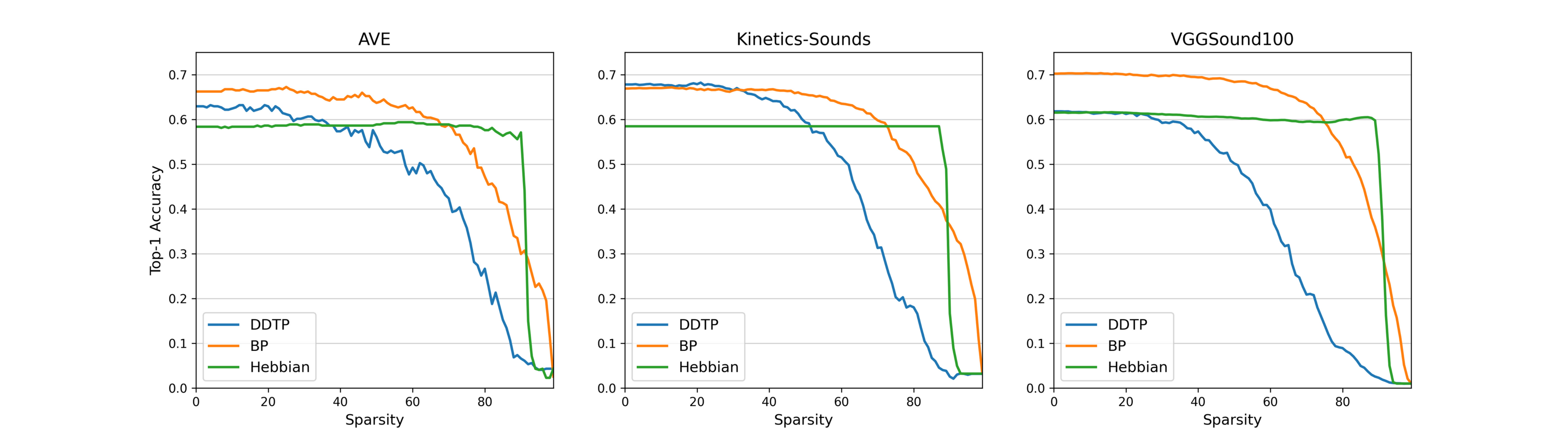}
    \caption{Top-1 accuracy after the Phase-2 VIB readout versus layer-wise magnitude sparsity for networks with five hidden layers trained with BP, DDTP, and the Hebbian rule. Each point corresponds to post hoc removal of the smallest absolute weights without retraining.}
    \label{fig:sparsity_ablation}
\end{figure*}

\noindent Sparsification affects Top-1 accuracy differently across learning rules and datasets. In deep architectures, Hebbian-trained networks generally show smaller descriptive reductions in Top-1 accuracy under post hoc synaptic removal than BP and DDTP, although DDTP can achieve higher absolute Top-1 accuracy in Kinetics-Sounds and VGGSound100. These heterogeneous responses to synaptic removal suggest that the underlying representations differ in how task-relevant information is distributed across connections. In particular, they motivate a post hoc sensitivity analysis of whether performance depends primarily on a small subset of high-magnitude synapses or on more broadly distributed connectivity patterns across layers.

We investigated this question through ablation studies based on layer-wise magnitude pruning. The networks were trained to convergence for 500 epochs and then pruned iteratively. For each sparsity level \(s \in \{1\%,2\%,\dots,99\%\}\), the \(s\%\) of synapses with the smallest absolute weights were removed independently in each layer, and Top-1 accuracy was evaluated without retraining. This dense-to-sparse protocol conceptually parallels activity-dependent pruning observed in cortical circuits \cite{braitenberg1998cortex, chechik1998synaptic}.

As shown in Fig.~\ref{fig:sparsity_ablation}, DDTP and BP exhibit measurable accuracy degradation as sparsity increases, with a pronounced drop in the high-sparsity regime when pruning is applied without retraining. By contrast, the Hebbian-trained networks are less sensitive to additional post hoc magnitude pruning in this pruning sweep. Accuracy remains relatively stable across a broad sparsity range and deteriorates only when connectivity is reduced below \(\sim 10\%\). This behavior follows from the effective magnitude sparsity induced by constrained Hebbian dynamics. Consistent with previous observations for the same learning principle \cite{neucomp}, the combination of nonnegative weights and competition among postsynaptic units produces a skewed weight distribution in which many synapses have zero or negligible magnitude. Consequently, additional layer-wise magnitude pruning of up to \(80\text{--}85\%\) primarily removes weights with little functional contribution to the Phase-2 readout. Top-1 accuracy degrades markedly only once pruning affects the remaining high-magnitude support of the learned representation, i.e., when connectivity falls below \(\sim 10\%\). The pruning sweep therefore identifies the sparsity regime in which pruning begins to affect the functional support of the representation, rather than reflecting an additional post hoc optimization or retraining effect. We note that this ultra-sparse regime lies below typical order-of-magnitude estimates of cortical synaptic connectivity \cite{braitenberg1998cortex,chechik1998synaptic}. Given the biological motivation for the sparsity constraint, performance degradation in this regime is not a central focus of this analysis.

\subsubsection{Ablation Studies and Literature Comparison}

\noindent This section reports unimodal ablations, additional control experiments, and literature-based reference comparisons. To assess the contribution of each modality, we performed ablation studies under unimodal conditions, proportionally reducing layer widths to reflect the decreased input dimensionality (Section~\ref{architecture}). The relative performance trends across datasets and methods mirror those observed in the bimodal setting, although absolute Top-1 accuracy is lower for single-modality inputs, as expected. These results are reported in Table~\ref{res2}.

\begin{table}[H]
\begin{center}
\centering
\caption{Ablation and control experiments. Top-1 accuracy after the Phase-2 VIB readout and task-information cost (CTI, unitless) are reported as mean \(\pm\) standard deviation over five independent seeds, unless otherwise indicated. The MNIST experiments evaluate learning without pretrained audiovisual embeddings. The VGGSound100 (VS100) controls compare the constrained Hebbian model against classical subspace-learning baselines, ablated variants, and a 10-hidden-layer depth-scaling control.}
\resizebox{\textwidth}{!}{%
\begin{tabular}{l c c c cc}
\hline
\textbf{Method} & \textbf{Dataset} & \textbf{Sparse} & \textbf{Beta (\(\beta\))} & \textbf{Top-1 [\%]} & CTI\\
\hline
BP & MNIST & Yes & \(1 \times 10^{-2}\) & \(96.5 \pm 1.6\) & \(9.3 \pm 0.3\) \\
BP (nonneg.) & MNIST & Yes & \(1 \times 10^{-2}\) & \(92.0 \pm 0.1\) & \(17.0 \pm 0.8\) \\
DDTP & MNIST & Yes & \(1 \times 10^{-2}\) & \(94.4 \pm 0.3\) & \(9.4 \pm 0.4\) \\
Hebbian (Ours) & MNIST & Yes & \(1 \times 10^{-2}\) & \(91.7 \pm 0.2\) & \(12.1 \pm 0.7\) \\
\hline
Generalized Hebbian Algorithm (GHA) \cite{sanger1989gha} & VS100 & No & \(1 \times 10^{-2}\) & \(62.0 \pm 0.3\) & \(38.6 \pm 1.5\) \\
Classical Oja subspace rule \cite{Oja1989} & VS100 & No & \(1 \times 10^{-2}\) & \(61.3 \pm 0.5\) & \(20.0 \pm 0.9\) \\
PCA-to-readout & VS100 & No & \(1 \times 10^{-2}\) & \(67.8 \pm 0.8\) & \(334.6 \pm 7.7\) \\
Constrained Hebbian without normalization/rescaling & VS100 & Yes & \(1 \times 10^{-2}\) & \(59.6 \pm 1.0\) & \(22.3 \pm 1.2\) \\
BP (nonneg.) with skip connections & VS100 & Yes & \(1 \times 10^{-2}\) & \(44.5^{\dagger}\) & \(20.8^{\dagger}\) \\
Depth-scaling control; Hebbian (Ours), 10 hidden layers & VS100 & Yes & \(1 \times 10^{-2}\) & \(7\text{--}10^{\ddagger}\) & -- \\
\hline
\end{tabular}%
}
\label{tab:ablations_controls}
\footnotesize{\(\dagger\)Only one valid run was available; no standard deviation is reported. \(\ddagger\)For the 10-hidden-layer depth-scaling control, no stable multi-seed estimate was obtained; the entry reports the observed approximate Top-1 range.}
\end{center}
\end{table}

\noindent Since we adopt the preprocessing routines of \cite{pian2023} for AVE, Kinetics-Sounds, and VGGSound100, their reported results provide natural reference benchmarks. Accuracy-based comparisons on these audiovisual datasets are scarce, as they are predominantly used to evaluate retrieval, captioning, or contrastive representation learning rather than strict classification performance. Following \cite{pian2023}, we use the performance ranges observed across multiple audio--visual class-incremental learning (CIL) baselines as reference points. These ranges, summarized in Table~1 of \cite{pian2023}, include LwF, iCaRL, SS-IL, and AFC, with accuracies spanning 42.4--74.04\% for AVE, 41.18--73.06\% for Kinetics-Sounds, and 26.21--72.8\% for VGGSound100 \cite{li2017lwf, rebuffi2017icarl, ahn2021ssil,kang2022afc}. \cite{pian2023} additionally report theoretical upper bounds obtained with full-dataset training of 76.85\%, 80.43\%, and 78.63\%, respectively. Complementary reference points are provided by \cite{wu2022large}, who report full-dataset accuracies on VGGSound ranging from 46.6\% to 75.4\% using large-scale contrastive pretraining.

Our results fall within these benchmark ranges, although they remain below the full-dataset upper bounds. This is consistent with the temporal aggregation of visual and audio embeddings, which reduces sequence-specific information, and with the additional constraints imposed by sparsification and local learning. The unimodal and bimodal trends are nevertheless preserved across datasets, indicating that the analyses capture stable modality-dependent structure rather than isolated readout effects.

Table~\ref{tab:ablations_controls} reports additional ablation and control experiments that separate the effects of pretrained upstream embeddings, classical subspace-learning rules, normalization and rescaling operations, and increased network depth. The MNIST control evaluates the same Phase-2 VIB readout pipeline using raw pixel inputs rather than pretrained audiovisual embeddings. This experiment tests whether compact, task-informative intermediate representations can also be obtained when the input is not already structured by a pretrained VideoMAE or AudioMAE encoder. BP and DDTP achieve the highest Top-1 accuracies in this setting (\(96.5\%\) and \(94.4\%\), respectively) and yield the lowest CTI values (\(9.3\) and \(9.4\), respectively). The constrained Hebbian model also forms a compact task-informative representation from raw pixels, reaching \(91.7\%\) Top-1 accuracy with a CTI of \(12.1\). Nonnegativity-constrained BP reaches comparable Top-1 accuracy (\(92.0\%\)) but with a higher CTI of \(17.0\). Thus, the MNIST control shows that the Hebbian rule can produce compressed, task-relevant representations without pretrained upstream feature extractors, while the main audiovisual experiments remain focused on downstream associative plasticity over fixed pretrained embeddings.

The VGGSound100 controls compare the constrained Hebbian setting with classical subspace-learning baselines and ablated variants. The Generalized Hebbian Algorithm (GHA) \cite{sanger1989gha} and classical Oja subspace rule \cite{Oja1989} achieve competitive Top-1 accuracy, indicating that classical subspace learning provides a strong unsupervised reference point for this task. The PCA-to-readout baseline applies an explicit linear PCA projection before the same downstream readout, without the constrained Hebbian hidden-layer learning rule. It therefore serves as a control for whether the observed performance can be explained by classical variance-preserving subspace compression alone. PCA-to-readout achieves the highest Top-1 accuracy among these controls, but at a much larger task-information cost. This indicates that higher predictive performance can be obtained by retaining substantially more representational information. The constrained Hebbian variant without normalization and activation rescaling was evaluated as a shallow-network ablation, because removing these stabilizing operations did not yield stable deeper-network estimates beyond one hidden layer, consistent with previous observations that constrained Hebbian dynamics become increasingly sensitive to depth in the absence of stabilizing normalization mechanisms \cite{neucomp}. In this shallow comparison, the ablated variant remains in a low-CTI regime but shows slightly reduced Top-1 accuracy. This suggests that the normalization and rescaling operations contribute to stable predictive performance under sparsity and nonnegativity constraints, rather than merely accounting for the low task-information cost.

The 10-hidden-layer depth-scaling control evaluates whether the constrained Hebbian setting extends to substantially deeper architectures. This configuration did not yield a stable multi-seed estimate and produced only low Top-1 performance. This result differs from prior work showing that related Hebbian mechanisms can scale reliably to 10 hidden layers when no explicit information-compression objective is imposed \cite{neucomp}. The present control therefore indicates that the combination of increased depth, stochastic encoder--decoder readout, and VIB-based bottleneck regularization can progressively remove task-relevant information before the readout. We interpret this as a limitation of the present compressed implementation and as evidence that substantially deeper constrained Hebbian networks require additional stabilization or architectural mechanisms when combined with explicit representational compression. Because the 10-hidden-layer configuration did not yield a stable VIB readout and remained only slightly above chance, CTI was not interpreted for this control.

\section{Discussion}

\noindent Our results demonstrate that, under the evaluated sparse structural constraints, Hebbian learning shifts the relationship between task-relevant information and representational cost toward a regime consistent with local synaptic resource allocation. This effect is observed across the AVE, Kinetics-Sounds, and VGGSound100 benchmarks, where Hebbian-trained networks exhibit lower CTI than sparse BP and DDTP in the main compressed comparisons, while reaching a CTI range comparable to shallow nonnegativity-constrained BP. In that sense, the audiovisual tasks function as probes of representational organization rather than as the primary object of optimization. In terms of Top-1 accuracy, the effect is not uniform across datasets or architectures. Hebbian performance is close to the sparse baselines on VGGSound100, exceeds sparse BP and DDTP in the deep AVE setting, and shows a mixed pattern on Kinetics-Sounds, where it remains competitive with sparse BP but does not match the strongest DDTP results. The nonnegativity-constrained BP control further indicates that low CTI can also be obtained by imposing a nonnegative constraint. However, this configuration remains below dense BP and DDTP in the shallow setting and collapses in the corresponding five-layer configurations. Overall, Hebbian plasticity shifts models toward a more favorable trade-off between CTI and functional accuracy.

Differences in the representational inductive biases of the learning rules, including their optimization objectives, locality assumptions, update dynamics, and induced distributions of weights and activations, provide a principled explanation for their different responses to enforced sparsification. Standard BP optimizes a global supervised objective and is not explicitly penalized for retaining input-dependent information that is not required for correct prediction. DDTP also preserves task-relevant information in several settings, but often at higher representational cost. By contrast, the constrained Hebbian rule is consistent with compact and decorrelated latent representations that allocate capacity preferentially to task-relevant dimensions. In a fixed network size, these distinctions contribute to dataset-specific outcomes. AVE appears comparatively less demanding relative to the model capacity, Kinetics-Sounds exhibits a favorable network--dataset capacity balance, and VGGSound100 remains the most demanding benchmark in the present evaluation.

Quantitatively, standard BP and DDTP preserve high \(I(Z;Y)\), but often at the cost of increased \(I(Z;H)\), particularly in deeper architectures. By contrast, the Hebbian rule typically maintains lower \(I(Z;H)\) while retaining task-relevant information, yielding favorable trade-offs between predictive information and representational cost in the evaluated shallow and five-layer settings. These gains also persist under enforced sparsification, indicating that constrained Hebbian learning can allocate representational capacity efficiently when architectural depth remains within the tested range.

At the same time, increasing network depth reduces Top-1 classification accuracy in several configurations, consistent with cumulative information loss in the stochastic encoder before the decoder readout \cite{silva2025, achille2018}. This effect is particularly relevant under the VIB-based compression objective used here. Prior work showed that related Hebbian mechanisms can scale reliably to 10 hidden layers in the absence of explicit information compression \cite{neucomp}. However, the present depth-scaling control indicates that the combination of substantially increased depth and compression introduces additional instability and loss of task-relevant information. Thus, the observed efficiency of Hebbian learning should be interpreted as a property of the constrained compressed regime evaluated here, rather than as evidence for unrestricted depth scalability. Overall, Hebbian learning preserves task-relevant information at lower CTI than BP and DDTP in the main evaluated settings, while substantially deeper compressed architectures require additional stabilization or architectural mechanisms.

Mechanistically, the low task-information costs of Hebbian-trained networks are consistent with a layer-wise redistribution of representational variance. Successive layers progressively decorrelate their outputs. At the same time, they concentrate representational capacity onto task-relevant dimensions, effectively implementing a local PCA-like decomposition of the input distribution. Early layers capture the dominant modes of the input, whereas subsequent layers encode residual variance not already represented, producing compact, task-focused latent codes throughout the hierarchy.

This structure tends to limit retained information about the input while preserving task-relevant information over the evaluated bottleneck settings, consistent with the empirical trends in Fig.~\ref{fig:deep}. Stochastic encoding and information bottlenecks further constrain the incremental information each layer can convey. Within the evaluated shallow and five-layer settings, the resulting decorrelation is associated with lower proxy-estimated representational cost. However, the 10-hidden-layer control indicates that this effect does not automatically extend to substantially deeper compressed architectures. Previous work confirms that Hebbian-induced structuring can reduce \(I(Z;H)\) even in shallow networks such as MNIST, demonstrating that compression can arise from layer-wise PCA-like organization itself rather than only from additional stochastic or architectural constraints \cite{bwhpc}.

This layer-wise redistribution of representational capacity produces a local organization in which task-relevant information is concentrated onto a limited subset of neurons and synapses, yielding sparse, low-redundancy representations. Conceptually, this mirrors neuronal competition during memory allocation, where eligibility and intrinsic excitability determine which synapses represent a specific memory \cite{josselyn2018, rao2019}. Through correlation-based, per-synapse updates, the Hebbian rule intrinsically implements this form of local resource allocation without access to global connectivity, supervision, or retraining.

Functionally, the pruning sweeps in Fig.~\ref{fig:sparsity_ablation} indicate that, without retraining, Hebbian-trained networks are less sensitive to post hoc weight removal than BP and DDTP in the high-sparsity regime. This pattern is consistent with the interpretation that BP and DDTP depend more strongly on globally coordinated synaptic configurations to preserve predictive performance under weight sparsification. In contrast, the Hebbian rule provides a direct local mechanism for coupling task-relevant information \(I(Z;Y)\) with an efficient and locally assigned set of connections. This distinction supports the use of \(I(Z;H)\) as a biologically motivated proxy for the representational burden associated with maintaining functional synaptic states.

Taken together, these results indicate that Hebbian learning generates sparse, task-relevant representations with lower representational cost through local, correlation-driven updates. While this highlights the rule's capacity for biologically plausible resource allocation, its assumptions and implementation choices require validation in future experiments. In the present implementation, we use an excitatory Hebbian update with an Oja-like population-wide competitive correction term as defined in Equation~\eqref{local}. This correction term sums across postsynaptic units and therefore imposes an apparent symmetry that may not be available to individual cortical neurons. Preliminary pilot investigations in which divisive normalization was restricted to local neighborhoods of postsynaptic units, thus approximating the information available to single neurons \cite{carandini}, indicated that locality-restricted normalization preserves the principal benefits of the Hebbian update while reducing the reliance on strict global symmetry. These observations are consistent with the cortical instantiation discussed in Section~\ref{rule} and with the theoretical considerations in \cite{friston}.

Recent work provides a complementary mechanistic perspective on this symmetry by deriving a formally equivalent population-corrected update
\begin{equation}
\Delta D_{ij} \propto z_j \Big(x_i - \sum_k D_{ik} z_k \Big),
\end{equation}
\noindent as the exact gradient of an efficient-coding objective in spiking networks \cite{Mikulasch2021}. The key insight is that the apparently global correction term can, in principle, be made locally available. Recurrent inhibition and, in compartmentalized neuron models, dendritic voltages can reflect an error-like residual that gates synaptic change through voltage-dependent plasticity \cite{Mikulasch2021,Brendel2020,Clopath2010}. Complementary work by the same authors recasts this idea in a local information-theoretic framework, in which neurons optimize decomposed objectives that explicitly regulate retained information and redundancy \cite{Makkeh2025}. Together, these studies demonstrate that learning rules of this mathematical form can produce efficient, compact, and nonredundant representations while remaining compatible with local neurobiological mechanisms.

Local realizability is supported in theory, and under the current implementation, these rules can be applied to deeper networks without compromising training stability, although additional layers do not yield gains in predictive accuracy. Mechanistically, this behavior arises from two interacting factors. First, enforcing Dale's law via strict nonnegativity of synaptic weights defines the set of admissible input–output transformations, constraining network expressiveness. Second, competitive Hebbian updates operate as successive PCA-like projections that predominantly transmit residual variance to subsequent layers. Together, these mechanisms produce a plateau in predictive performance for deeper architectures, reflecting current implementation design choices. Previous work has shown that applying a similar Hebbian rule without enforcing strict nonnegativity increases depth scalability on standard visual benchmarks (e.g., MNIST), indicating that the excitatory-only constraint, rather than the learning rule itself, accounts for the observed performance plateau \cite{b2}. A natural next step, with benefits beyond increasing biological plausibility, is to introduce explicit inhibitory populations following canonical cortical proportions of 80\% excitatory neurons and 20\% inhibitory neurons \cite{rubenstein2003, gatto2010}. This modification can increase network expressivity and allow normalization and error-like signals to arise through local excitatory–inhibitory interactions, potentially obviating the batch-wise z-normalization currently required in excitatory-only implementations. 

Another avenue for future work could involve integrating local Hebbian updates with feedback-driven allocation of representational resources or synaptic and neuronal tagging mechanisms \cite{frey1997, redondo2011}. While inhibitory populations may alleviate expressivity constraints, these complementary mechanisms could further refine hierarchical and class-specific representations across layers \cite{felleman1991}. Systematic evaluation across network depth, sparsity, and multimodal tasks will be necessary to determine how these interventions, in combination with excitatory-inhibitory network architectures, affect classification performance, representational diversity, and representational efficiency.

Future work could also explore applying the proposed local Hebbian learning rule to alternative architectures, such as convolutional networks (CNNs), to assess the generality of its representational and resource-allocation properties. While our study focused on mechanistic evaluation in a controlled MLP setting, convolutional layers could potentially enhance predictive performance. However, many Hebbian CNN implementations, including FastHebb \cite{fasthebb}, NM-Hebb \cite{mili_NMHebb}, and related models \cite{Miconi2021_HebbianGradients, journe2022hebbian}, rely on strong architectural priors such as exact translational weight sharing or hybrid gradient-based fine-tuning, which are not directly supported by neuroanatomical evidence \cite{pogodin2021towards}. For example, in our replication of FastHebb, pretraining alone achieves only $\sim$35\% accuracy on CIFAR‑10, whereas a subsequent gradient-based fine-tuning phase increases performance to $\sim$85\% \footnote{Original FastHebb code: \url{https://github.com/GabrieleLagani/HebbianLearning}.}. Similarly, NM-Hebb employs metric-learning fine-tuning, and \cite{pogodin2021towards} implement sleep-like consolidation with lateral weight equalization while still relying on gradient updates during the wake phase. These cases illustrate that strictly local Hebbian updates, without any gradient-based or supervisory intervention, remain difficult to scale to deep convolutional architectures while preserving biological fidelity. More broadly, recent self-supervised representation-learning methods, sparse-training algorithms, and modern pruning strategies may yield different accuracy--compression trade-offs, but they introduce additional objectives, architectural priors, retraining schedules, or explicit parameter-selection mechanisms that are not controlled in the present study. The observed reduction in CTI should therefore be interpreted within the evaluated BP, DDTP, and constrained Hebbian baseline set, and as complementary to the broader space of sparse, pruned, and self-supervised representation-learning methods.

Finally, the metabolic cost metric used here, approximated by mutual information \(I(Z;H)\), provides a relative measure of coding efficiency rather than a direct biophysical energy estimate. Its biological interpretation relies on the theoretical assumption that reduced representational complexity is associated with a reduced burden on the underlying biophysical states that instantiate the representation. This information-theoretic proxy therefore neglects contributors to neuronal energy consumption such as spike-timing variability, membrane currents, and glial support \cite{attwell2001energy, sterling2015principles}. Consequently, \(I(Z;H)\) and CTI should be interpreted as comparative indicators of representational burden, not as direct measurements of metabolic expenditure. Future studies should incorporate experimentally calibrated energy metrics, such as ATP consumption, electrophysiological energy budgets, or metabolic imaging data.

It is important to note that the present findings focus on synaptic resource allocation in higher-level representational regimes. The fixed VideoMAE and AudioMAE embeddings hold early sensory processing constant, thereby isolating downstream associative plasticity. The added raw-input MNIST control indicates that the evaluated learning rules can also produce task-informative representations without pretrained audiovisual encoders. Nevertheless, the main audiovisual results should be interpreted as evidence for efficient allocation over fixed upstream representations, rather than as a full account of end-to-end feature discovery from raw pixels or waveforms. Future studies could investigate whether similar plasticity mechanisms retain these properties when applied to fully unprocessed bimodal sensory inputs.

\subsection{Conclusion}

\noindent In this study, we used audiovisual classification as a controlled assay to evaluate whether local, structurally constrained Hebbian plasticity can support efficient synaptic resource allocation in higher-level representational regimes. Under constraints inspired by anatomical structure and metabolic considerations, standard BP achieves strong predictive performance. Biologically inspired alternatives such as DDTP achieve comparable accuracy, but neither is routinely assessed with respect to biologically grounded resource constraints. In contrast, the excitatory Hebbian rule implements key neurobiological principles, including sparsity, locality, and Dale's law, while producing task-relevant representations with lower proxy-estimated representational cost. Our framework extends beyond performance benchmarking by formalizing a tractable metric, grounded in the FEP and VIB, for comparing algorithms in terms of relative representational cost under a biologically motivated information-theoretic proxy. This metric is intended as a comparative measure of resource allocation across learning rules, not as a direct estimate of biological energy expenditure.

Future work could examine whether the same resource-allocation properties extend beyond fixed audiovisual embeddings to fully unprocessed multimodal inputs and to non-audiovisual domains such as text, graphs, and structured tabular data. Another avenue for future work could be to assess whether the low-CTI representations induced by constrained Hebbian learning are also more selective, disentangled, or interpretable than BP- or DDTP-trained representations, for example using unit-level selectivity measures, representational-similarity analyses, disentanglement metrics, and attribution-based comparisons. Future work could also apply this framework to recurrent, spiking, and inhibitory--excitatory architectures, as well as other advanced network designs, including feedback-modulated plasticity, to explore systematic comparisons across contemporary learning algorithms. Assessing the framework on neuromorphic hardware that implements Hebbian learning \cite{Kim2025_HebbianNeuromorphic, Zhou2025_NeuromorphicHebbianMTJ} would provide further insight into its applicability and mechanistic relevance across both algorithmic and hardware substrates.

\end{document}